\documentclass[9pt]{article}
\usepackage[utf8]{inputenc}
\usepackage{INTERSPEECH2022,amsmath,graphicx}
\usepackage{multirow}
\usepackage{textcomp}
\usepackage{hyperref}
\hypersetup{colorlinks=true}
\usepackage{moredefs}
\usepackage[final]{changes}

\usepackage{bibspacing}
\usepackage{color}
\definecolor{darkspringgreen}{rgb}{0.39, 0.65, 0.27}

\begin{document}
\ninept

\title{Improving Deliberation by Text-Only and Semi-Supervised Training}
\name{Ke Hu, Tara N. Sainath, Yanzhang He, Rohit Prabhavalkar, Trevor Strohman, Sepand Mavandadi, Weiran Wang}
\address{Google LLC, USA}
\email{huk@google.com}

\maketitle
\vspace{-0.5em}

\begin{abstract}
Text-only and semi-supervised training based on audio-only data has gained popularity recently due to the wide availability of unlabeled text and speech data. In this work, we propose incorporating text-only and semi-supervised training into an attention-based deliberation model. By incorporating text-only data in training a bidirectional encoder representation from transformer (BERT) for the deliberation text encoder, and large-scale text-to-speech and audio-only utterances using joint acoustic and text decoder (JATD) and semi-supervised training, we achieved 4\%-12\% WER reduction for various tasks compared to the baseline deliberation. Compared to a state-of-the-art language model (LM) rescoring method, the deliberation model reduces the Google Voice Search WER by 11\% relative. We show that the deliberation model also achieves a positive human side-by-side evaluation compared to the state-of-the-art LM rescorer with reasonable endpointer latencies.
\end{abstract}
\vspace{-0.2em}

\section{Introduction}

End-to-end (E2E) automatic speech recognition (ASR) models have made tremendous improvements in recent years~\cite{sainath2021efficient,  narayanan2021cascaded, chen2021developing, li2020developing, yeh2019transformer, wang2021cascade, saon2021advancing, li2021recent}. In a state-of-the-art system~\cite{sainath2021efficient}, a neural language model (LM) is used to rescore a cascaded encoder model and outperforms a conventional  ASR system  in  both  Google Voice Search (VS)  and  rare word recognition quality, as well as latency. The LM in~\cite{sainath2021efficient} is trained using billions of text-only data and proves to improve rare word recognition quality. \added{While LM relies on only text hypotheses for rescoring, deliberation models have been recently proposed for second-pass rescoring using both text hypotheses and audio ~\cite{hu2020deliberation, hu2021transformer}. Compared to LM training, there has been few attempts at incorporating widely available text-only or audio-only data in deliberation (see ~\cite{mavandadi2021deliberation}). In this work, we research various ways to utilize large-scale text-only and semi-supervised data for deliberation training.}

While the LM in~\cite{sainath2021efficient} uses causal conformer layers, bidirectional textual context is incorporated by using bidirectional encoder representations from transformers (BERT)~\cite{shin2019effective, fohr2021bert}. In addition to LM rescoring, neural correction models train text-to-text models to predict targets based on estimated transcripts. For example, a BERT model is used in~\cite{hrinchuk2020correction} to initialize a transformer neural correction model. To increase diversity, text-to-speech (TTS) utterances are decoded to generate text hypotheses to train a transformer correction model~\cite{li2020developing}. LSTM models are used similarly in~\cite{guo2019spelling}. However, since neural correction only relies on text, its correction capability is potentially limited and thus only used for spelling correction.

Instead of training external modules such as LMs, several recent studies incorporate text-only data into supervised training to jointly train E2E models~\cite{tang2021general, mavandadi2021deliberation, deng2021improving, bapna2021slam, renduchintala2018multi, bahar2019comparative}. For example, text-only data has been used to train speech encoders~\cite{tang2021general, bapna2021slam}. \cite{tang2021general} leverages text-only data represented as phonemes and masked by noise, and uses them as inputs to predict the corresponding text  using a shared encoder with ASR. In~\cite{bapna2021slam}, either text-only or speech-only data have been used in a speech-text joint training to pre-train an encoder for a downstream task such as ASR. On the other hand, ASR decoders have also been modified for text-only training. \cite{mavandadi2021deliberation} extends a joint acoustic and text decoder (JATD) from the Listen, Attend and Spell (LAS)~\cite{sainath2020attention} to a deliberation decoder, and uses text-only data  (or synthesized utterances) to train the decoder with fixed context vectors. \cite{deng2021improving} modifies the transformer decoder to have only self-attention (except for the last layer) so they can be trained by text-only data.

Besides text-only data, audio-only data is also widely available and thus used to assist ASR training~\cite{schneider2019wav2vec, baevski2020wav2vec, zhang2021bigssl, doutre2021improving, hwang2021large}. For example, by pre-training speech encoders using audio-only data, wav2vec~\cite{schneider2019wav2vec, baevski2020wav2vec} achieves competitive results by using a small amount of labeled data. In~\cite{zhang2021bigssl}, the authors use large-size models up to billions of parameters in semi-supervised learning using unlabeled data combined with a small portion of labeled data. The idea of noisy student training is explored in~\cite{doutre2021improving, hwang2021large}, where a bidirectional teacher generates training data for a streaming student by using noisy corrupted inputs.

In this work, we propose to incorporate text-only data to pre-training the deliberation text encoder in a masked language model (MLM) task similar to BERT~\cite{devlin2018bert}. Our results show that pretraining a conformer text encoder with large enough size significantly improves recognition for both Voice Search and long-tail words. In addition to the text encoder, we also synthesize large-scale text-only data (84M) to TTS utterances in training the deliberation decoder using JATD~\cite{mavandadi2021deliberation}.
Third, since deliberation attends to encoded audio, we perform large-scale semi-supervised training using 500M unlabeled speech utterances from Google Voice Search domain and transcribed using a conventional model. With all the proposed techniques, we achieved 4\%-12\% WER reductions for various test sets compared to the deliberation baseline. Compared to a state-of-the-art LM rescorer~\cite{sainath2021efficient}, our deliberation model performs 11\% relative better in VS, 16\% for the SxS test set, and competitively for long-tail. A human side-by-side comparison shows the deliberation performs significantly better than LM rescoring with reasonable endpointing latencies.
\vspace{-0.5em}

\section{Modeling Improvement}

\subsection{Model Overview}
\begin{figure}[h]
  \centering
   \includegraphics[scale=0.4]{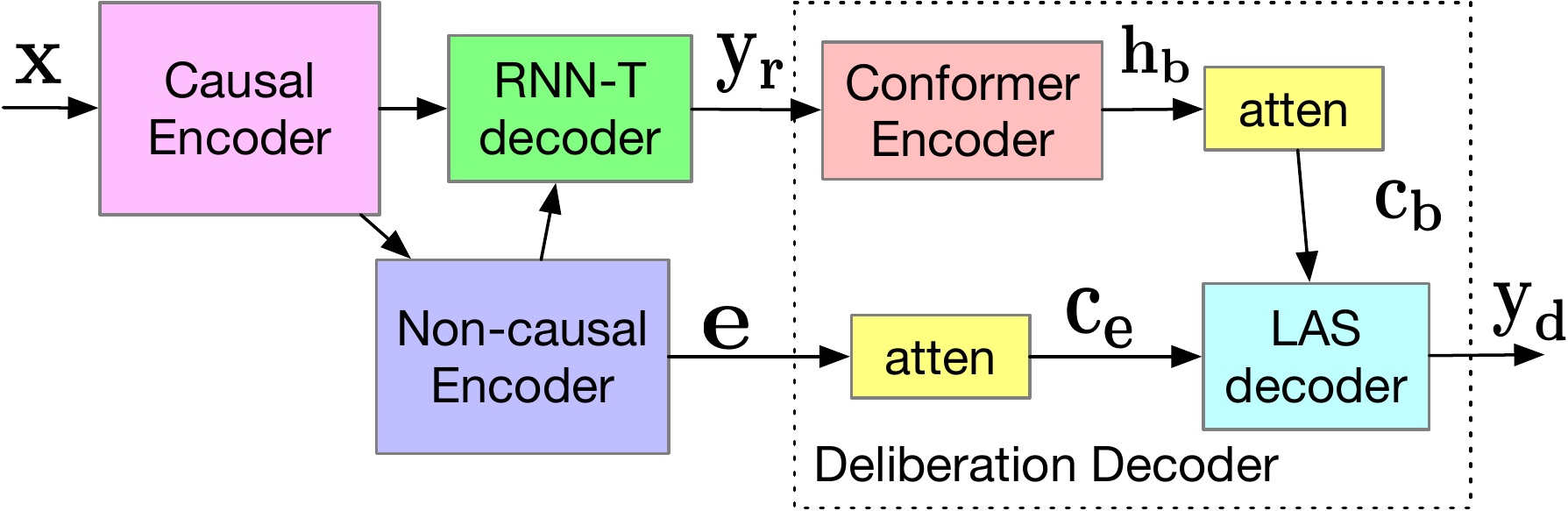}
   \caption{Deliberation based on cascaded encoders.}
   \label{fig:delib}
   \vspace{-2em}
\end{figure}

Our model is illustrated in Fig. \ref{fig:delib}. Note that different from~\cite{hu2020deliberation, hu2021transformer}, the deliberation decoder is based on the non-causal encoder~\cite{narayanan2021cascaded} instead of a causal encoder. The decoder attends to both the non-causal encoder output ($\bf{e}$) and hypotheses ($\bf{y_r}$) from the non-causal path, i.e., decoded using non-causal encoder. The non-causal encoder often has a right-context for better recognition quality~\cite{narayanan2021cascaded}. We use a conformer encoder~\cite{gulati2020conformer} as the text encoder. A two-source attention LAS decoder is used as the deliberation decoder, similar to~\cite{hu2020deliberation}. The decoder can be used for either re-decoding or rescoring. The deliberation model in this work does not stream compared to~\cite{hu2022transducer}, as we focus on text-only and semi-supervised training.
\vspace{-1em}

\subsection{BERT Training Based on Text-Only Data}

\subsubsection{Pretrained BERT Text Encoder}
\label{sec:bert}

The conformer encoder in deliberation in Fig. \ref{fig:delib} is a text encoder which takes text hypotheses ($\bf{y_r}$) as inputs and outputs text encodings ($\bf{h_b}$). In regular deliberation training~\cite{hu2020deliberation, hu2021transformer}, we randomly initialize the parameters of the text encoder and train it jointly with other parts of the deliberation model. The training uses only supervised data. In this work, to leverage large amounts of text-only data, we propose to pretrain the text encoder alone by masking the text-only inputs and then predicting the masked tokens using a cross entropy (CE) loss, similar to BERT~\cite{devlin2018bert}.

Given an input text sequence, we first tokenize it into wordpieces, and randomly choose 15\% of tokens, similar to~\cite{devlin2018bert}. Each token is then replaced with \texttt{[MASK]} for 80\% of the time, a random token for 10\% of the time, or kept unchanged for 10\% of the time. Since our input hypotheses are single sentences, we use only one segment for each BERT input. We use \texttt{<sos>} to replace the special classification token \texttt{[CLS]}, and use \texttt{<eos>} for \texttt{[SEP]}. The \texttt{<sos>} and \texttt{<eos>} share the same vocabulary as the cascaded encoder model to match BERT inputs to the format of first-pass hypotheses. Each input sequence is padded to a fixed length, and padded tokens are not used for computing the CE loss. We reuse the token id of \texttt{<epsilon>} for \texttt{[MASK]} since it does not appear in our hypotheses which contain only non-blank labels. Lastly, the task of next sentence prediction is removed from our task since we only have single text sentences. Instead of transformer layers, we use conformer layers~\cite{gulati2020conformer} for the text encoder. We refer to a text encoder trained in this way as a pretrained BERT.

The pretrained BERT is then used as the deliberation text encoder, and we update the parameters of the BERT and deliberation decoder jointly in training. The cascaded encoder is kept frozen. We have also tried freezing the entire or part of the pretrained BERT (e.g. layers close to inputs) and update the rest of the deliberation decoder, but this leads to worse quality.

\subsubsection{Masking First-Pass Hypotheses}
Similar to the MLM idea in BERT~\cite{devlin2018bert}, we also try to increase the diversity of the text encoder inputs by randomly masking tokens in the input hypotheses. This aims to increase the diversity of the data seen by the text encoder. Note that this is different from the pretraining in Sect. \ref{sec:bert} because here we do not use any external data to pre-train the text encoder. We predict the text targets use the CE loss in training, and do not use any masking in inference. We show in Sect. \ref{sec:ablation_mlm} that simply masking a small portion of first-pass hypothesis tokens in training improves over the baseline deliberation in certain conditions.

\subsection{Large-Scale TTS and Semi-Supervised Training}

\subsubsection{Large Scale TTS Training}
\label{sec:jatd}

Text-only data can also be converted to TTS utterances for training the whole deliberation decoder. We employ JATD training~\cite{mavandadi2021deliberation} and scale it up using text data sampled from multiple domains, i.e., 51M,  20M,  1.6M, 0.6M, and 11M text sentences from Maps, News, Play, Search and YouTube domains, respectively. In comparison,~\cite{mavandadi2021deliberation} uses only 4.6M samples from the Maps domain. The text sentences are then converted to speech by using a multi-speaker TTS system~\cite{gonzalvo2016recent}. In our large scale training, we mix the TTS data and supervised training data in a 1:9 ratio. The deliberation decoder is trained from scratch using the mixed data. We find that sampling data using probabilities proportional to their amount gives the best result. To differentiate supervised and TTS data, we use only data from a single domain for each training batch. When the data is supervised, we compute both audio and text attention during training, and otherwise use fixed context vectors to replace both audio and text attention for TTS data. We show that Maps-only data improves the corresponding domain quality, but degrade other domains, resulting similar average performances. However, by using data from all domains, JATD significantly improves all long-tail sets on average as well as the Maps domain (details in Sect. \ref{sec:ablation_jatd}).

\subsubsection{Large Scale Semi-Supervised Training}
\label{sec:semi}

Apart from text-only data, we also explore using large-scale unlabeled audio utterances to improve training, inspired by~\cite{zhang2021bigssl, doutre2021improving}.
In total, we have 500M audio-only utterances sampled from the Search domain and then generate estimated transcripts using a state-of-the-art hybrid model~\cite{pundak2016lower} for training.
The model employs a state-of-the-art language model for decoding and tends to generates quality training data for long-tail words.
The utterances are not augmented by any noise.
\deleted{We have tried adding the multi-condition noise~\cite{kim2017generation} to speech utterances but that leads to regression. This is probably because added noise causes the quality of estimated transcripts to drop.} \added{Note that in later experiments (Sect. \ref{sec:overall_improve}), we also mix the large-scale TTS data (84M) and the semi-supervised data (500M) in deliberation training. As far as we know this is the first attempt to experiment with large-scale TTS and semi-supervised training for deliberation. In training, we use the TTS utterances as 10\% of all data, semi-supervised data as 10\%, and train the deliberation model from scratch.}

\vspace{-0.5em}
\section{Experiments}
\label{sec:exp}

We perform our experiments using large-scale data~\cite{narayanan2019recognizing} based on a state-of-the-art cascaded encoder model~\cite{li2021better}.

\subsection{Modeling Details}

\subsubsection{Baseline Deliberation Model}
Our deliberation model is based on a cascaded encoder baseline~\cite{li2021better} which consists of 17 causal conformer layers and 5 non-causal layers. Each causal layer has a model dimension of 512 with 8-headed self-attention. 
The five non-causal layers has a total of right context of 0.9s.
We use an embedding prediction network as in~\cite{botros2021tied}.
The cascaded encoder model is trained to predict 4,096 lowercase wordpieces~\cite{schuster2012japanese}.

Our LAS-based deliberation decoder attends to non-causal encoder outputs and hypotheses decoded using the non-causal encoder.
For efficiency, we use 4 first-pass hypotheses. The text encoder is a 2-layer 640-D conformer encoder with a two-token right-context, totaling around 12M parameters. We use 8-headed attention for both audio and hypotheses. The deliberation decoder consists of 2 LSTM layers (similar to~\cite{hu2020deliberation}), where each layer has 2,048 hidden units followed by 640-dimensional projection. A 4,096-dimensional softmax is then used to predict the same wordpieces as the baseline cascaded encoder. The decoder has around 42M parameters.

An input speech waveform is divided into 32-ms segments using hanning windows at a rate of 10 ms to compute 128-D log-Mel features. Each log-Mel feature is then stacked with three previous frames to form a 512-D vector, which is then downsampled to a 30-ms frame rate as input features.

\subsubsection{Training Data}
\label{sec:train_data}

Our multi-domain (MD) supervised training data consists of around 300M utterances described in~\cite{narayanan2019recognizing}. The utterances are sampled from multiple domains, and are anonymized and hand-transcribed except for YouTube where utterances are generated using a semi-supervised method~\cite{liao2013large}. In total, we have \texttildelow400k hours of training data. We also increase the data diversity by using multi-condition training~\cite{kim2017generation} such that the utterance signal-to-noise ratio (SNR) is between 0dB and 30dB. We also use mixed-bandwidth utterances at 8kHz or 16 kHz~\cite{yu2013feature}, and SpecAug~\cite{park2019specaugment}.
For text-only training, we use a text corpus which contains more than 100B sentences~\cite{sainath2021efficient} to train the BERT text encoder described in Sect. \ref{sec:bert}. The text-only data spans multiple domains including  Maps,  News,  Play,  Search  and YouTube.

\subsubsection{Test Data}
\label{sec:test_data}

We use three sets for evaluation. The Voice Search (VS) test set contains \texttildelow14K anonymized and hand-transcribed utterances sampled from general Google Voice Search traffic. A SxS test set contains around 900 utterances where an E2E model~\cite{sainath2019twopass} performs inferior to a state-of-the-art hybrid model~\cite{pundak2016lower}. To focus on long-tail word recognition, we use a long-tail (LT) test set described in~\cite{peyser2020improving}. The LT utterances are synthesized using text sentences containing words rare in multi-domain training, or with surprising pronunciations~\cite{peyser2020improving}.
The utterances range across multiple domains such as Maps, News, Play, Search and YouTube, totaling 200K.
In the following ablation studies, we use a subset of LT, called Rare Proper Noun Maps (RPNM), to represent the LT set for experiment efficiency. The performance of RPNM usually correlates well with the full LT set.

\subsection{Ablation Studies}
\label{sec:ablation}

\subsubsection{Pretrained BERT Text Encoder}

We compare three pretrained BERT (PTB) text encoders, described in Sect. \ref{sec:bert}, with large, medium and small sizes. The PTBs have 12, 4, and 2 conformer layers with a model dimension of 512, corresponding to 76M, 32M, and 12M parameters, respectively. The conformer right context is set to 30 tokens to become ``bidirectional". We use the same wordpiece model as the deliberation for tokenization. We can see in Table \ref{tab:bert} that the large PTB performs the best, with a VS WER of 4.6\%. The WER improvement is uniform for all test sets, ranging from 4\% to 12\% compared to the baseline deliberation. We note that when the size of the PTB reduces, the improvement reduces gradually. When using a small PTB with the same size as the baseline deliberation, we did not see any benefits from pretraining. This indicates that a relatively large BERT may be needed to leverage the large amount of unpaired text data.

\begin{table}[h]
\caption{WERs (\%) of deliberation using pretrained and non-pretrained BERT text encoders of different sizes.}
\vspace{-1em}
\begin{tabular}{ |l|c|c|c|c| }
    \hline
    \multirow{2}{*}{Model} & \# Text Enc. & \multicolumn{3}{|c|}{WER (\%)} \\ \cline{3-5}
    &  Layers  & VS & SxS & RPNM \\ \hline
    Deliberation & 2L & 4.8 & 26.9 & 12.0 \\ \hline
    \hspace{0.02in} + Large PTB & \multirow{2}{*}{12L} & \bf{4.6} & \bf{23.6} & \bf{11.1} \\ \cline{1-1} \cline{3-5}
    \hspace{0.06in} - PT & & 5.3 & 29 & 13.2 \\ \hline
    \hspace{0.02in} + Medium PTB & \multirow{2}{*}{4L} & 4.7 & 25.4 & 11.5 \\ \cline{1-1} \cline{3-5}
    \hspace{0.06in} - PT & & 4.8 & 26.5 & 12.0 \\ \hline
    \hspace{0.02in} + Small PTB & \multirow{2}{*}{2L} & 4.8 & 26.3 & 11.9 \\ \cline{1-1} \cline{3-5}
    \hspace{0.06in} - PT & & 4.8 & 26.3 & 11.9 \\ \hline
\end{tabular}
\label{tab:bert}
\vspace{-2em}
\end{table}

To further analyze whether the improvement is due to increased size of the BERT, we remove pretraining (PT) for all PTBs. We see in Table \ref{tab:bert} that there is significant regression in large and medium PTB scenarios for all test sets. In the large BERT scenario, we also note that the model becomes hard to train without text-only pretraining, resulting worse performance than the medium-size BERT. In addition, the small non-pretrained BERT performs similar to the baseline, indicating our small conformer BERT has a similar performance as the original conformer encoder. In addition, we have also tried even larger BERT with 16 and 24 layers but none of them obtained uniform improvements for all test sets compared to the 12L BERT. Considering computation efficiency, we choose the 12L BERT as the text encoder for the following experiments.

\subsubsection{Semi-Supervised Training}

As described in Sect. \ref{sec:semi}, we use 500M semi-supervised utterances with transcripts generated using a hybrid model~\cite{pundak2016lower} for training. We mix the semi-supervised data with the supervised data using a 1:9 ratio and train the deliberation model from scratch. The semi-supervised data mainly improves VS WER by 4\% relatively (4.6\% $\rightarrow$ 4.4\% in Table \ref{tab:semi}). This is probably because our unlabeled data is from the Voice Search domain. In this study, our semi-supervised data size is relatively small compared to text-only data, in future we plan to generate more semi-supervised data using more powerful teachers.

\begin{table}[h]
\vspace{-1em}
\caption{WERs (\%) by semi-supervised training.}
\vspace{-1em}
\begin{tabular}{ |l|c|c|c|c| }
    \hline
    \multirow{2}{*}{Model} & \# Text Enc. & \multicolumn{3}{|c|}{WER (\%)} \\ \cline{3-5}
    & Layers &  VS & SxS & RPNM \\ \hline
    Large PTB & \multirow{2}{*}{12L} & 4.6 & 23.6 & 11.1 \\ \cline{1-1} \cline{3-5}
    \hspace{0.02in} + Semi-sup. data & & \bf{4.4} & \bf{23.6} & \bf{10.9} \\ \hline
\end{tabular}
\label{tab:semi}
\vspace{-2em}
\end{table}

\subsubsection{Rescoring}

So far, our decoding is done by beam search. To compare to LM rescoring later (Sect. \ref{sec:compare}), we use deliberation to rescore the first-pass hypotheses in a teacher-forcing fashion~\cite{sainath2019twopass}. Compared to Table \ref{tab:semi}, the rescoring WERs are 4.6\%, 25.5\%, and 10.9\%, for VS, SxS, and RPNM, respectively. This is expected according to our previous findings~\cite{hu2020deliberation, hu2021transformer}. We use the deliberation rescorer for later experiments.

\subsubsection{JATD}
\label{sec:ablation_jatd}

In Table \ref{tab:jatd}, we see that by using only Maps data (JATD-Maps), JATD improve significantly on the RPNM test set, which is a set focusing on Maps. But there is no improvement in LT, indicating potential regression for other domains. We thus perform large-scale JATD training for all domains described in Sec. \ref{sec:train_data} (JATD-All). Table \ref{tab:jatd} shows that we have achieved significant improvements on both RPNM and LT. The improvement is around 9\% relative for the LT set. Note that our deliberation rescoring baseline here does not incorporate pretrained BERT. Similar to~\cite{mavandadi2021deliberation}, we notice VS and SxS results do not change significantly.

\begin{table}[h]
\vspace{-1em}
\caption{WERs (\%) by rescoring using JATD training.}
\vspace{-1em}
\begin{tabular}{ |l|c|c|c|c|c| }
    \hline
    \multirow{2}{*}{Model} & \# Text Enc. & \multicolumn{4}{|c|}{WER (\%)} \\ \cline{3-6}
    & Layers &  VS & SxS & RPNM & LT \\ \hline
    Delib. Rescoring & \multirow{3}{*}{2L} & 4.9 & 27.6 & 12.0 & 30.7 \\ \cline{1-1} \cline{3-6}
    \hspace{0.02in} + JATD-Maps & & 5.0 & 26.8 & 10.4 & 30.7 \\ \cline{1-1} \cline{3-6}
    \hspace{0.02in} + JATD-All & & 5.0 & 26.9 & \bf{10.3} & \bf{27.8} \\ \hline
\end{tabular}
\label{tab:jatd}
\vspace{-2em}
\end{table}

\subsubsection{Apply Masking to Hypotheses}
\label{sec:ablation_mlm}

To increase the diversity of data for text encoder training, we have also tried applying masking to first-pass hypotheses, similar to the MLM idea in~\cite{devlin2018bert}. Specifically, we mask around 2\% of hypothesis tokens, randomize only 0.01\% tokens, and leave the rest unchanged. We have tried other ratios but did not find improvement. We experiment masking to three deliberation models in Table \ref{tab:mlm}. Overall, we notice that masking only improves the vanilla deliberation rescoring by 3.6\% relative for the SxS set. When other techniques such as pretrained BERT or JATD are used, masking degrades the SxS and LT performance. We thus will not include this in our final system but recommend as a convenient approach to increase data diversity for the baseline deliberation.

\begin{table}[h]
\vspace{-0.5em}
\caption{WERs (\%) using masking for first-pass hypotheses.}
\vspace{-1em}
\begin{tabular}{ |l|c|c|c|c| }
    \hline
    \multirow{2}{*}{Model} & \# Text Enc. & \multicolumn{3}{|c|}{WER (\%)} \\ \cline{3-5}
    & Layers &  VS & SxS & LT \\ \hline
    Delib. Rescoring & \multirow{2}{*}{2L} & 4.9 & 27.6 & 30.7  \\ \cline{1-1} \cline{3-5}
    \hspace{0.02in} + MLM & & 5.0 & \bf{26.6} & 30.6 \\ \cline{1-3} \cline{3-5}
    Delib. + PTB & \multirow{2}{*}{12L} & 4.8 & 24.9 & 29.5 \\ \cline{1-1} \cline{3-5}
    \hspace{0.02in} + MLM & & 4.8 & 25.8 & 29.7 \\ \hline
    Delib. + JATD-All & \multirow{2}{*}{2L} &
    5.0 & 26.9 & 27.8 \\ \cline{1-1} \cline{3-5}
    \hspace{0.02in} + MLM & & 5.0 & 28.0 & 27.9 \\ \hline
\end{tabular}
\label{tab:mlm}
\vspace{-2em}
\end{table}

\section{Comparison}
\subsection{WER Comparisons}
\label{sec:overall_improve}

In Table \ref{tab:wer_compare}, we compare the cascaded encoder baseline (\texttt{B0}) to the baseline deliberation model (\texttt{B1}) and deliberation with proposed training techniques (\texttt{E1}-\texttt{E3}). The cascaded encoder model (\texttt{B0}) is exactly the first-pass model used for deliberation.

First, we see that our best-performing deliberation model (\texttt{E3}), with all techniques proposed in the paper, performs significantly better than the deliberation baseline (\texttt{B1}), reducing WERs by 4.1\%, 6.5\%, and 11.7\%, for VS, SxS, and LT test sets, respectively. The improvement is more prominent for long-tail sets, indicating the text-only and semi-supervised training is especially effective for long-tail. Compared to cascaded encoder (\texttt{B0}), our WER improvement is up to 15\%. For individual techniques, we see that JATD and semi-supervised training improves LT significantly by around 10\% relative.
BERT pretraining improves VS significantly (6\% relative), and the lack of LT improvement is probably because JATD already does well in long-tail.

In Table \ref{tab:wer_compare}, we also compare to a LM rescoring model (\texttt{B2}) similar to~\cite{sainath2021efficient}, which consists of 12 conformer layers. The LM has a model dimension of 384 and 3072-D feedforward layers, 4-headed self attention, and a left context of 31 tokens. Overall, the LM rescorer has 71M parameters. The LM is trained using the same text-only data used to train the BERT text encoder. During inference, the conformer LM is used to rescore the lattice after the non-causal cascaded encoders.
Compared to LM rescoring in Table \ref{tab:wer_compare}, deliberation with JATD (\texttt{E1}) performs similarly to LM rescoring (\texttt{B2}) in long-tail words, and 6\% and 12\% relatively better for VS and SxS test sets, respectively. Note that without BERT encoder the deliberation rescorer size of \texttt{E1} is 57M, which is 20\% smaller than LM (71M). When incorporating BERT and semi-supervised training, we achieve more significant and uniform improvements: VS (8.9\%), long-tail (1.8\%), and SxS test set (15.6\%), all in relative WER reductions. 

\begin{table}[h]
\vspace{-0.5em}
\centering
\caption{WER (\%) improvements by deliberation rescoring using text-only and semi-supervised training.}
\vspace{-1em}
\begin{tabular}{ |c|l|c|c|c|c| }
    \hline
    \multicolumn{2}{|c|}{\multirow{2}{*}{Model}} & \# Text Enc. & \multicolumn{3}{|c|}{WER (\%)} \\ \cline{4-6}
    \multicolumn{2}{|c|}{} & layer & VS & SxS & LT \\ \hline
    B0 & Cas. Enc.~\cite{narayanan2021cascaded} & - & 5.4 & 30.2 & 31.6  \\ \hline
    B1 & Deliberation & \multirow{2}{*}{2L} & 4.9 & 27.6 & 30.7  \\ \cline{1-2} \cline{4-6}
    E1 & Delib. JATD-All & & 5.0 & 26.9 & 27.8 \\ \cline{1-4} \cline{4-6}
    E2 & \hspace{0.02in} + 12L-BERT & \multirow{2}{*}{12L} & 4.7 & 25.9 & 27.9 \\ \cline{1-2} \cline{4-6}
    E3 & \hspace{0.02in} + Semi-sup & & \bf{4.7} & \bf{25.8} & \bf{27.1} \\ \hline
    B2 & LM Rescoring & 12L & 5.3 & 30.6 & 27.6 \\ \hline
\end{tabular}
\label{tab:wer_compare}
\vspace{-2em}
\end{table}

\subsection{Side-by-Side Comparison with LM Rescoring}
\label{sec:compare}

We further compare the proposed deliberation model to a state-of-the-art LM rescorer~\cite{sainath2021improving} in a decoding setup with endpointing. The baseline cascaded encoders in~\cite{sainath2021improving} consist of a small causal encoder and large non-causal encoder. 
The causal encoder consists of a 7-layer conformer and the non-causal encoder has a 10-layer right-context conformer with an overall right-context of 0.9s. The encoder output dimension is projected to 384 to reduce model size.
Following~\cite{sainath2021efficient}, we use the hybrid autoregressive transducer (HAT) version of the LM rescoring model (the non-HAT version performs worse). For deliberation, we take the best-performing rescorer (\texttt{E3} in Table \ref{tab:wer_compare}) and reduce the decoder dimension to 384 to match the cascaded encoder. The deliberation rescorer has a total size of 106M. We found that a non-HAT decoder works better for deliberation than the HAT version.

We compare deliberation and LM rescoring in a human side-by-side evaluation. A total of 705 utterances are transcribed by both models, and are sent to two human transcribers to rate. Each transcript is rated as either a win for deliberation over LM rescoring (only deliberation is correct), or a loss (only LM rescoring is correct), or neutral (both models are correct or incorrect).
Table \ref{tab:sxs_epon} shows the deliberation rescorer changes 9\% of traffic, and has significantly more wins (114) than losses (50) compared to LM rescoring. Overall, the p-Value of $<0.1\%$ shows the difference is statistically significant.

\begin{table}[h]
  \vspace{-1em}
  \centering
  \caption{Side-by-side eval: LM vs. Delib. Rescoring}
  \vspace{-1em}
  \begin{tabular}{|c|c|c|c|c|} \hline
    Changed (\%) & Win & Loss & Neutral & p-Value \\ \hline
    9.0 & 114 & 50 & 541 & $<$0.1\%  \\ \hline
  \end{tabular}
  \label{tab:sxs_epon}
  \vspace{-1em}
\end{table}

In terms of VS WERs in this comparison, the deliberation model achieves a WER of 5.4\%. This is 10\% relative better than cascaded encoders (6.0\%), and 8\% better than HAT LM rescoring (5.9\%). We have also tried increasing the LM size to around 100M \added{or using a BERT LM} but did not see any improvement. The deliberation model achieves an EP50 (median latency) of 380 ms and EP90 (90th latency) of 720 ms, similar to LM rescoring.

\vspace{-1em}
\section{Conclusion}
\label{sec:conclude}

We researched text-only and semi-supervised training for LAS-based deliberation. By incorporating pretrained BERT text encoder, large-scale JATD and semi-supervised training, we have improved the deliberation performance by 4\% for VS, and 12\% relative for long-tail in terms of WERs. In the latest cascaded encoder setup with endponting, we show the proposed deliberation rescorer outperforms a state-of-the-art LM rescoring method by 8\% relative in terms of VS WER, and wins in a human side-by-side evaluation.

\bibliographystyle{IEEEtran}
\bibliography{refs}

\end{document}